# Evaluation of geographical distortions in language models

**Rémy Decoupes[1,2] · Roberto Interdonato[1,3] · Mathieu Roche[1,3] · Maguelonne Teisseire[1,2] · Sarah Valentin[1,3]**

**Abstract**
Geographic bias in language models (LMs) is an underexplored dimension of model fairness, despite growing attention being given to other social biases. We investigate whether LMs provide equally accurate representations across all global regions and propose a benchmark of four indicators to detect undertrained and underperforming areas: (i) indirect assessment of geographic training data coverage via tokenizer analysis, (ii) evaluation of basic geographic knowledge, (iii) detection of geographic distortions, and (iv) visualization of performance disparities through maps. Applying this framework to ten widely used encoder- and decoder-based models, we find systematic overrepresentation of Western countries and consistent underrepresentation of several African, Eastern European, and Middle Eastern regions, leading to measurable performance gaps. We further analyse the impact of these biases on downstream tasks, particularly in crisis response, and show that regions most vulnerable to natural disasters are often those with poorer LM coverage. Our findings underscore the need for geographically balanced LMs to ensure equitable and effective global applications.

## 1 Introduction

Nowadays, language models (LMs) are widely used as sources of information across a variety of applications, from search and retrieval to question answering and text analysis. Owing to their ability to encode and retrieve knowledge effectively, they are increasingly comple-

✉ Rémy Decoupes
remy.decoupes@inrae.fr

1 TETIS, Univ. Montpellier, AgroParisTech, CIRAD, CNRS, INRAE, Maison de la Télédetection, 500, rue J.F.Breton, 34090 Montpellier, France

2 INRAE, Montpellier, France

3 CIRAD, Montpellier, France

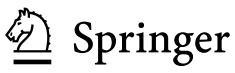

menting or replacing traditional tools such as search engines and encyclopaedic resources such as Wikipedia. These models can be understood as compressed representations of large volumes of textual data from the internet, a substantial portion of which contains a spatial dimension (Manvi et al., 2023).

Moreover, questions related to geography, travel, and cultural aspects represent the third most important use of LMs, after those about programming and artificial intelligence (Zheng et al., 2023). In addition, some kinds of information, such as spatially disaggregated indicators, are not directly accessible through Earth observations and spatial imagery (Manvi et al., 2023), although they are sometimes available in LM knowledgebases. These findings reinforce the use of LMs for questions related to geography and downstream applications. For instance, the spatial information included in LMs can be beneficial for social-good-related AI applications, such as crisis management or humanitarian aid (Belliardo et al., 2023).

LMs, although built on a common architecture known as transformers (Vaswani et al., 2017), can be categorized into two distinct groups: masked language models (MLMs) and causal language models (CLMs). The first group, which is based solely on one part of the architecture, the encoder, has significantly advanced the field of natural language processing (NLP) across a range of tasks, including text classification and named entity recognition. These models are trained to predict masked words in sentences, which gives rise to their category name. BERT (Devlin et al., 2019) is a prominent example. CLMs, which constitute the second group, are based either on an encoder-decoder architecture or solely on a decoder. They are trained to predict the next word, enabling them to generate coherent and relevant text. These models are often referred to as large language models (LLMs), not only because of the size of their neural architectures but also because of the enormous scale of their training data. These models still contain the five sources of bias present in any NLP project (Hovy & Prabhumoye, 2021), namely, biases in the data, annotations, vector representations, models and research design. We define bias as a *systematic and repeatable error that creates unfair outcomes, such as privileging one arbitrary group of users over others.* Biases become problematic when they are propagated or amplified through model hallucinations (Wan et al., 2023). Hallucinations correspond to *plausible yet nonfactual content* (Huang et al., 2025) generated by an LM. Inherent biases in pretraining data can lead to inequalities in information representation and contribute to knowledge gaps, ultimately resulting in data-related hallucinations (Huang et al., 2025). Certain types of bias are particularly associated with hallucinations, notably those related to gender (Wan et al., 2023) and nationality (Venkit et al., 2023).

In this study, we focus on the biases inherent to spatial information representation. While social biases such as gender, sexual orientation, and ethnicity are often studied (Navigli et al., 2023), they rarely include biases related to geographical knowledge representation (Mai et al., 2023). These distortions in representation lead to decreased performance on downstream tasks (Decoupes et al., 2023).

The main goal of this paper is to provide an evaluation framework for both MLMs and CLMs to identify two biases: spatial disparity of geographic knowledge and geographic distortion, and compare them with the spatial distribution of locations in the training datasets. The evaluation is both reproducible[1] and scalable to all LMs. A short version of this work appeared in (Decoupes et al., 2025). As illustrated in Fig. 1, this framework is built upon four indicators: spatial information coverage in training datasets, spatial disparities in geographical knowledge, correlation between geographic and semantic distances and their outliers. These criteria are then employed to evaluate the ten most commonly utilized models. To clarify our approach, we propose the following definitions. Geographic knowledge refers to the models' knowledge of human geography. Conjoint analysis with the first two indicators reveals that, in MLMs, countries with the highest-quality geographic knowledge tend to be those most represented in the training datasets. Therefore, we refer to these as overrepresented countries, in contrast to underrepresented countries, which are those that are sparsely present in the training datasets. Furthermore, when semantic and geographic distances are not correlated, we refer to this as both distortion and misrepresentation of geographic distances.

[1] https://github.com/tetis-nlp/geographical-biases-in-llms

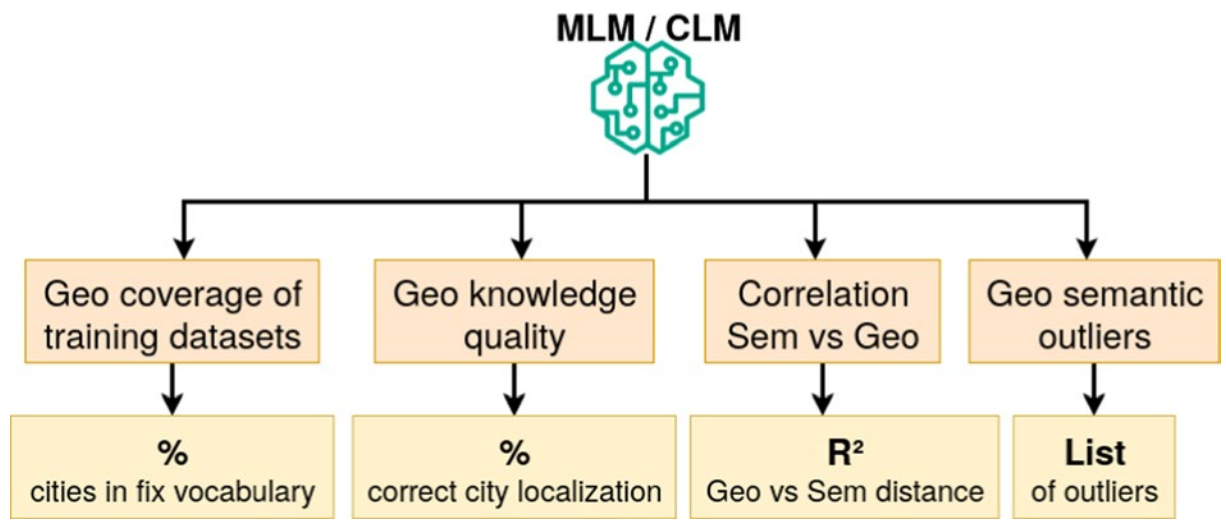

**Fig. 1** Methodology overview for geographic bias identification: the four indicators

We also assess the impact of these geographical biases on downstream NLP tasks, with a particular focus on using LMs for crisis management.

The rest of the paper is organized as follows. Section 2 outlines related work. Section 3 presents our contributions. Section 4 evaluates ten LMs, highlights geographical bias and proposes an analysis of the consequences of these regional disparities on downstream applications (details are presented in Appendix A). Finally, future directions for this research are summarized in Sect. 5.

## 2 Related work

Language models have increasingly been applied to tasks involving geography, making it among the most common use cases for modern chatbots, ranked just behind programming and IT-related queries (Zheng et al., 2023). These models have demonstrated strong capabilities in understanding and reasoning about the world (Roberts et al., 2023).

In particular, a growing body of research explores how both MLMs (Huang et al., 2022; Li et al., 2023) and CLMs (Hu et al., 2023; Zhang et al., 2023a; Manvi et al., 2023; Roberts et al., 2023) can support a range of geographic information system (GIS) tasks. These include extracting place names or toponyms (proper names of places) (Hu et al., 2023); querying geographic databases such as OpenStreetMap[2]; retrieving information about demographic or infrastructural variables (Manvi et al., 2023; Mai et al., 2023), such as population density, income levels, or the number of hospitals; and even identifying relevant satellite imagery for spatial analysis (Zhang et al., 2023a).

To have a better impact on downstream tasks, several studies aim to enhance the geographical knowledge of models by injecting spatial information into the questions (or prompts) addressed to them (Hu et al., 2023; Mai et al., 2023). For instance, they can be assisted by describing the locations for which they need to make predictions. For example, CLMs can better predict population densities when they are provided with a list of points of interest (POIs) (Manvi et al., 2023). Another way to incorporate geographical knowledge into models is by modifying the neural network architecture with a merging layer between the vector representations (embeddings) and geographical knowledge graphs, enabling the enhancement of representations (Huang et al., 2022; Li et al., 2023).

[2] ChatGeoPT: https://github.com/earth-genome/ChatGeoPT

However, the truthfulness of the spatial knowledge in CLMs is not uniform; for instance, GPT-4 shows imprecision when asked to provide coordinates or distances between sparsely populated cities (Roberts et al., 2023), thus directly impacting the performance of these models on geography-related NLP tasks, such as crucial tasks in humanitarian crisis response (Decoupes et al., 2023). Indeed, as noted by Zhang et al. (2023b), errors produced by CLMs can compound, leading to increasingly significant inaccuracies. These errors are often rooted in hallucinations, i.e., responses generated by CLMs that appear plausible but are in fact incorrect or unfounded (Huang et al., 2025). Another source of error is geographic bias, which refers to the unequal coverage or consideration of different regions, countries, or cultures. This imbalance is already present in foundational geographic information sources such as Wikipedia and OpenStreetMap and is reflected more broadly across the web. As a result, geographic coverage tends to be uneven and biased towards urban, affluent, and English-speaking regions (Hecht & Stephens, 2014). These geographic biases can propagate into downstream tasks, leading to better performance for wealthier countries and regions of the Global North (Stillman & Kruspe, 2025), lower ratings assigned to locations in the Global South (Manvi et al., 2023), and noticeable regional disparities in factual accuracy (Faisal & Anastasopoulos, 2023).

It is acknowledged in the NLP community that models, and even text corpora themselves, can capture spatial distances; it is indeed possible to infer the geographical distance between two places on the basis of their co-occurrence in large corpora (Louwerse & Zwaan, 2009). More recently, Gurnee and Tegmark (2023) demonstrated that embeddings also effectively capture geographic distance. We believe it is particularly relevant to investigate biases related to the representation of places in relation to their geographic distance to infer the context (co-occurrence-based or not) in which these places have been associated.

One potential approach to mitigating geographic biases in CLMs is to train them to engage in reasoning, specifically, to recognize when they lack sufficient knowledge to answer a question and to retrieve relevant information from reliable external sources. However, as shown by Momennejad et al. (2023), for simple planning tasks, CLMs tend to rely on memorization rather than reasoning. When prompted or guided to reason explicitly, they often generate invalid or incoherent trajectories. Memorization is defined as a concept situated between generalization and regurgitation and refers to the verbatim reproduction of memorized text fragments (Hartmann et al., 2023). It involves minimal abstraction or generalization beyond surface-level pattern recognition.

This motivates the development of a dedicated evaluation framework, presented in the next section, to identify the regions most affected by geographic bias, with a particular focus on spatial distance distortions.

## 3 Geographical knowledge quality indicators

To assess geographic bias, we propose four indicators (as illustrated in Fig. 1) with two main objectives. The primary goal is to assess the quality of geographical knowledge to identify potential regions of the world that are less well known to the models. The second objective is to assess whether the representations of places are spatially biased by examining the relationships between geographical and semantic distances.

As listed in Table 1, we include two language model families, i.e., MLMs (e.g., BERT) and CLMs (e.g., ChatGPT). For the MLMs, we include the most well-known models and their multilingual versions: BERT (Devlin et al., 2019), multilingual-BERT (m-BERT), RoBERTa (Liu et al., 2019), XLM-RoBERTa (Conneau et al., 2020) and a BERT-based geospatial understanding model, GeoLM (Li et al., 2023). With respect to CLMs, we compare six of them by including two among the most reused open source models[3]: LLaMa (LLaMa 2 and LLaMa 3) (Touvron et al., 2023; Grattafiori et al., 2024) & Mistral (v0.1 (Jiang et al., 2023) and v0.3[4]). OpenAI/ChatGPT (GPT−3.5-turbo) is utilized with prompts, while OpenAI/Ada is employed to retrieve embeddings because of the unavailability of ChatGPT embeddings.

As languages significantly contribute to culture and consequently to geographical representation, we include multilingual models such as XLM-RoBERTa and m-BERT in our comparisons. Even though CLMs can generate multilingual text, they are still primarily

**Table 1** Parameters of the language models included in the study

| Models | Name | Languages | Parameters (million) | Vocabulary Size |
|---|---|---|---|---|
| BERT | Bert-base-uncased | English | 110 | 30522 |
| m-BERT | Bert-base-multilingual-uncased | 102 | 168 | 105879 |
| GeoLM | Geolm-base-cased | NA | NA | 28996 |
| RoBERTa | Roberta-base | English | 125 | 50265 |
| XLM-RoBERTa | xlm-roberta-base | 100 | 279 | 250002 |
| LLaMa 2 | Llama-2-7b-chat-hf | English | 7000 | 32000 |
| LLaMa 3 | Meta-Llama-3-8B-Instruct | English | 8000 | 128256 |
| Mistral-v01 | Mistral-7B-Instruct-v0.1 | English | 7000 | 32000 |
| Mistral-v03 | Mistral-7B-v0.3 | English | 7000 | 32768 |
| Ada | Text-embedding-ada-002 | NA | NA | 100277 |
| ChatGPT | gpt-3.5-turbo | NA | NA | 100277 |

NA (Not Available) is used when the parameters cannot be found in the associated paper trained on English sources. For example, LLaMa 2's pretraining data consists of approximately 89.7% English text (Touvron et al., 2023), and LLaMa 3 is trained on approximately 95% English data.[5] The exact language distribution for Mistral V0.1 and V0.3 has not been disclosed, although they are generally understood to have been trained predominantly on English sources. Concerning ChatGPT, we are also not aware of the language distribution during its training. Therefore, we cannot assert whether it is a multilingual model or predominantly English. For the ground truth, we use geographical data sourced from the GeoNames gazetteer extracted by OpenDatasoft, comprising cities with a population of more than 1000 inhabitants.[6] GeoNames is a geographical database that stores coordinates, elevation, population and administrative subdivisions for 25 million toponyms (proper names of locations or places).

[3] https://huggingface.co/spaces/HuggingFaceH4/open_llm_leaderboard

[4] https://huggingface.co/mistralai/Mistral-7B-v0.3

### 3.1 Disparity of geographical knowledge quality indicators

Two indicators are proposed to assess the disparity of geographical knowledge in language models across all regions of the world. By geographical knowledge, we refer to the model's ability to provide knowledge about human geography, such as political boundaries.

#### 3.1.1 Indicator 1: spatial information coverage in training datasets

The main source of bias comes from the quality of the training datasets (Navigli et al., 2023). Therefore, for this indicator, we propose indirectly assessing the spatial coverage of the training datasets. To do this, we examine the models' vocabulary, which corresponds to the most frequent words encountered during their pretraining. We therefore look at the number of cities by continent that appear in the list of words most frequently encountered.

Consequently, this experiment seeks to identify city names present in the vocabularies of language models, indirectly gauging the over- or underrepresentation of these cities by continent in the training datasets. We investigate the inclusion or exclusion of cities (using their English names) with a population of more than 100,000, totalling 4,916 cities, in the predefined vocabularies of the models. For example, London and Paris are in the BERT vocabulary, but Ouagadougou (the capital of Burkina Faso) is not.

#### 3.1.2 Indicator 2: Spatial disparities in geographical knowledge

This indicator aims to assess whether the geographical knowledge of the models is of equal quality across the globe. To achieve this, we use a geo-guessing task, where models are required to determine the location of a given place based on limited information, such as an image, description, or coordinates (Liu et al., 2024; Roberts et al., 2023). For the experiments in this article, the models must guess the country when provided with its capital or cities with populations over 100,000.

To calculate it, we had to adapt the probe to the two families of models (MLMs and CLMs). MLMs were initially pretrained to predict a masked word in a sentence. Based on the context of the masked word, these models can guess it (illustrated in Listing 1).

[5] https://ai.meta.com/blog/meta-llama-3/

[6] https://public.opendatasoft.com/explore/dataset/geonames-all-cities-with-a-population-1000

Given that CLMs are generative models, we propose querying them using natural language prompts, as illustrated in Listing 2.

```
masked_sentence_capital = f'{city} is capital of <mask>'.
masked_sentence_cities_100K =  f'The city of {city} is located in the country of <mask>'.
```

**Listing 1** Predicting masked country for encoder-based models

```
# Capital
{"role": "user", "content": "Name the country corresponding to its capital: Paris. Only give the country."},
{"role": "assistant", "content": "France"},
{"role": "user", "content": f"Name the country corresponding to its capital: {city}. Only give the country."}

# Cities with > 100K inhabitants
{"role": "user", "content": "Name the country containing Montpellier. Only give the country."},
{"role": "assistant", "content": "France"},
{"role": "user", "content": f"Name the country containing {city}. Only give the country."}]
```

**Listing 2** Question answering for LLMs

### 3.2 Geographical distance distortion indicators

In this section, we present two indicators that assess the correlation between geographic distances and semantic distances for pairs of locations. The models convert the words they encounter into high-dimensional vectors (768 for BERT and 4096 for LLaMa 2). These representations, also known as embeddings, help capture the semantics of words. For cities that are not part of the fixed vocabulary, we compute the average of the embeddings of all their subtokens.

According to Tobler (1970), on page 236, the first law of geography states that *Everything is related to everything else, but near things are more related than distant things*. This implies that the embeddings of geographically close cities should also be close to the semantic space of the models.

The embedding of a word is partly formed during the training phase by co-occurrence with other words encountered in the training set and in the inference phase, during which the context of its sentence modifies the pretrained embedding through the attention mechanism. According to Gurnee and Tegmark (2023), these embeddings also contain spatial information. Indeed, by training a small model (multilayer perceptron), they manage to correctly predict the GPS coordinates of locations from their embeddings. We rely on these results to introduce the next two indicators.

#### 3.2.1 Indicator 3: correlation between geographic distance and semantic distance

This indicator aims to analyse whether semantic representations (embeddings) are correlated with geographical distance. The objective is to assess, continent by continent, whether semantic representations consider the geographical distance between pairs of cities. Geographical distance represents the direct distance between two cities as the crow flies. Among all the semantic distances used in NLP (Azarpanah & Farhadloo, 2021), we opted for the one based on cosine similarity, as it is the most widely used. Cosine similarity measures the angle between two vectors while being invariant to their magnitude. Although alternative metrics such as Euclidean or Manhattan distance can also be used, they are more sensitive to differences in vector norm than to directional similarity. In high-dimensional latent spaces (e.g., BERT with 768 dimensions), these norm-sensitive distances tend to lose discriminative power, as many vectors appear equally distant from one another. In contrast, cosine similarity remains a more reliable and meaningful measure of semantic relatedness, as it focuses on the orientation of vectors rather than their length. The semantic distance $D_{\mathrm{sem}}$ is defined as the complement of the cosine similarity between the vectors corresponding to the cities in the embedding space:

$$D_{\mathrm{sem}} = 1 - \mathrm{cosine_similarity}(\mathrm{Embedding}_{\mathrm{city}_1}, \mathrm{Embedding}_{\mathrm{city}_2}) \tag{1}$$

### 3.2.2 Indicator 4: anomaly between geographical distance and semantic distance

This indicator aims to identify regions semantically distant from the rest of the world in terms of vector average, which could be attributed to an underrepresentation of these regions in the training data. To visualize these differences, we propose focusing only on the three largest cities by population in each country and calculating the average semantic distances separating them from all the other cities under consideration.

We therefore apply this indicator to all countries. To better measure semantic isolation, we also correct it for geographical isolation. For example, the average semantic distances from Sydney, Australia, to other cities in the world need to be corrected to compare them with the average semantics of a city in Europe since Australia is a geographically distant country. Therefore, we introduce the geographical distortion index (GDI), defined as follows:

$$GDI = \frac{1 + D_{sem}}{1 + \overline{D_{geo}}} \tag{2}$$

where $D_{sem}$ is the semantic distance between pairs of cities and $\overline{D_{geo}}$ is the normalized geographic distance among all the cities $\in [0, 1]$. One is added to the numerator and denominator for cases where the geographical or semantic distances are close to zero. Without this +1, the GDI would be close to 0 or infinity and would not be representative of extreme distortion. For example, let us consider two cities whose semantic distance is zero and geographical distance is very high. In this case, the value of GDI without +1 in the numerator would be close to 0, which would suggest that there is no distortion. This adjustment constrains the GDI values to between [0.5, 2]. Thus, when the GDI is close to 1, there is no distortion. If it is less than 1, the semantic distance is not significant enough. Conversely, if $GDI > 1$, the semantic distance is too high.

To illustrate the isolation or overrepresentation of certain areas of the world, we calculate, for each city with a population of 100,000, the five other cities that are closest semantically. Next, we compute two ratios: the percentage of closest semantic cities that are in the same country and the percentage that are on the same continent.

## 4 Results and implications

In this section, we present the most significant findings revealed by our four indicators and their impacts on downstream applications. The complete set of detailed results for each indicator is available in Appendix A.

### 4.1 Overview of key results

The global results at the continental scale for each model under comparison are presented in Fig. 2. Four main points should be highlighted. First, a high number of parameters does not guarantee that the model has good geographical knowledge. As shown in Fig. 2b, the BERT model (0.10 billion parameters) occasionally outperforms Mistral (7 billion parameters) on the geo-guessing task, indicating that the performance does not scale straightforwardly with model size. Furthermore, because the training datasets for these models are not publicly available, we cannot assess whether dataset size is correlated with performance.

The second point is in line with the fact that training datasets are the source of the strongest biases (Navigli et al., 2023). Indeed, for MLMs, countries that are underrepresented in the datasets (Fig. 2a) show lower performance in terms of other indicators. This suggests that such underrepresentation impacts the quality of predictions. These two observations open up a new perspective of research, i.e., to analyse the impact of training datasets on the number of parameters and assess the risks of diluting geographical information in favour of better performance according to other criteria. This is exemplified by the case of Mistral, which outperforms LLaMa-2-7B and even LLaMa-2-13B in most evaluations (Jiang et al., 2023) but predicts fewer countries given its capital than BERT and LLaMa-7-B. It is interesting to analyse when geographical information disappears in favour of other types of knowledge during the training process.

The third point is that multilanguage models partially correct the bias of the data by having more diversified training datasets that provide other cultural and geographical contexts; they perform better for African or Asian countries but at the expense of English-speaking countries. However, the extended training of multilingual models can reduce their ability to respond correctly to more complex prompts (Choshen et al., 2022), leading to poorer results, as illustrated by indicator 2 when applied to cities with 100,000 inhabitants, in comparison with the capital (Appendix A.2). On complex prompts, MLM (BERT and RoBERTa) perform better than their multilingually fine-tuned counterparts. Therefore, caution is needed when choosing to work with multilingual models on complex tasks in English. While multilingual models appear to have better geographic knowledge, they seem to lose response capability.

Finally, the last point is that countries in Oceania, although geographically isolated, appear to be paradoxically semantically close to other countries in the world, while Eastern Europe, Africa and Southeast Asia (Indonesia, Vietnam, and Cambodia) are semantically isolated.

This diversity in knowledge could impact NLP tasks related to information retrieval, event detection, and tracking. For example, in crisis management, identifying events (e.g., the nature of the event and location) is crucial. However, given the massive flow of information exchange, especially through social networks, it is essential to deduplicate events. An event can be described differently, and it is important not to interpret it as several separate

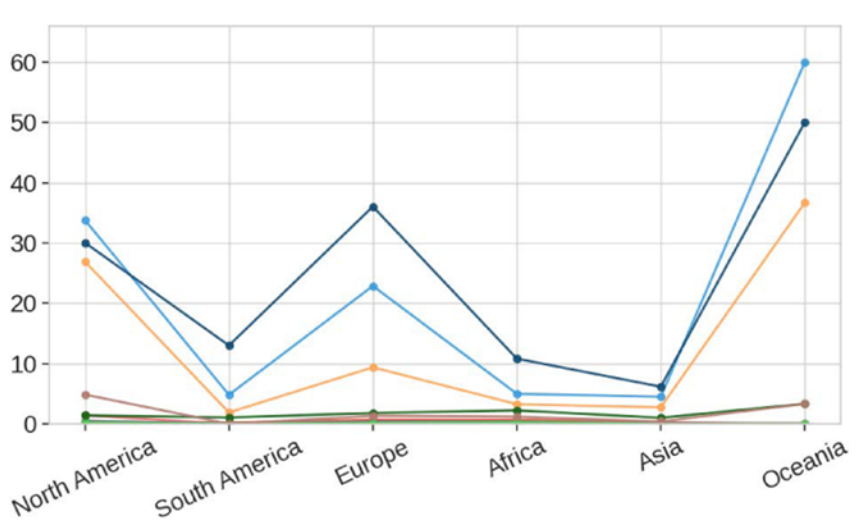

(a) Ind1: Geographical coverage of the training dataset

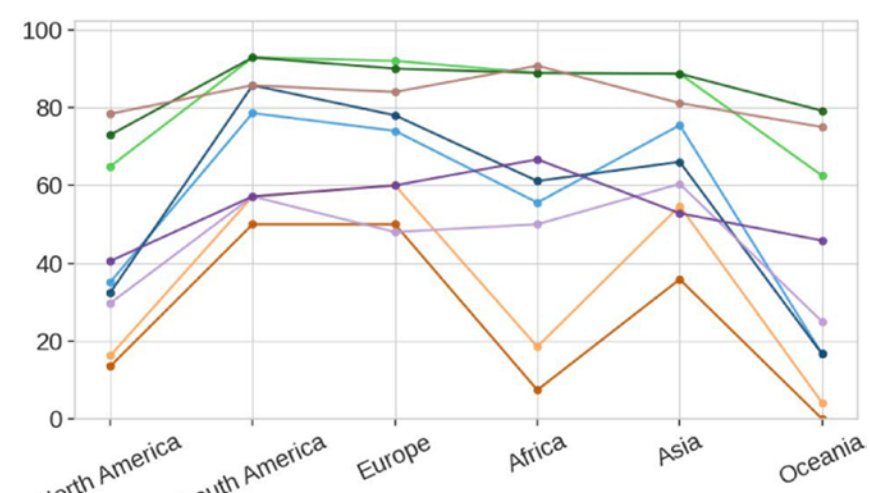

(b) Ind2: Geographical knowledge quality

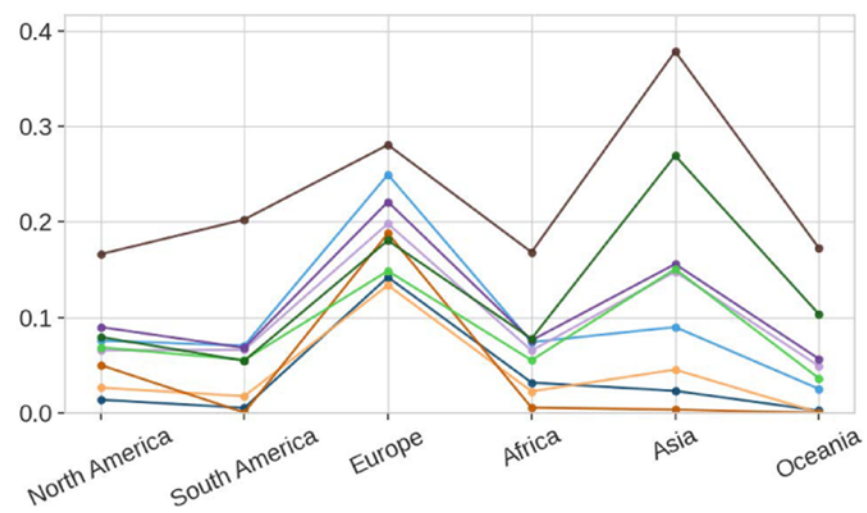

(c) Ind3: Correlation between geographical and semantic distances

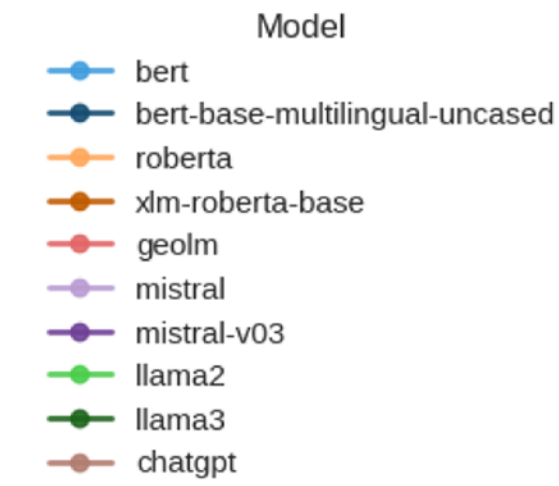

(d) Legend for the first three indicators

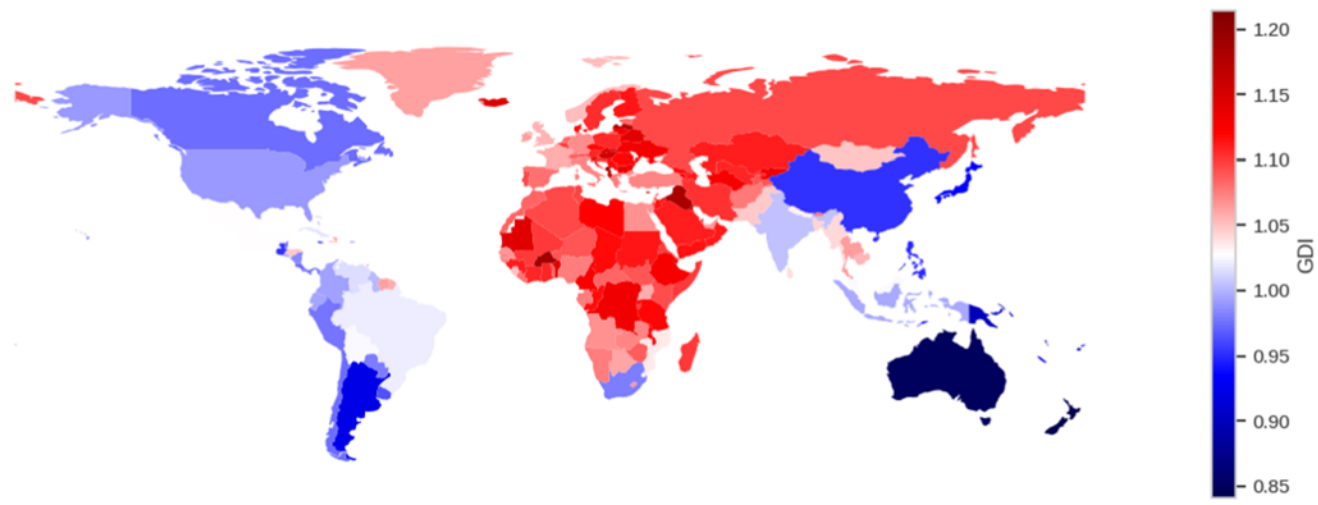

(e) Example of a result for Ind4: average per country of the BERT semantic distances from its three most populous cities to the cities of the world events. However, poor semantic representation of locations and distances leads to incorrect deduplication. Therefore, the contribution of these models may have a limited impact on aiding responses to humanitarian crises in countries that may need it the most. This point is discussed in detail in the next Sect. 4.2.

**Fig. 2** Overview of the results for the four indicators

Another perspective of research is how to improve the correlations between geographical distance (a proxy for cultural context) and semantic distance. How does improving this correlation lead to better performance in downstream tasks such as information retrieval and event detection and tracking? Does improving the quality of geographical knowledge lead to better results in geographical NLP tasks for location detection or GPS coordinate assignment?

### 4.2 Implications for downstream applications

This section seeks to identify the impact of geographical variability in model knowledge on downstream tasks. Typical natural language processing (NLP) tasks include text classification, named entity recognition, and conversational agents (chatbots).

However, measuring the influence of these geographical disparities presents challenges because the current evaluation methods commonly used in NLP provide scores only at the benchmark level. Indeed, the authors of (Liu et al., 2022) state that there are no existing datasets specifically designed to assess an NLP task on the basis of location found within phrases or authors' geographic origins. In that study, the researchers concentrate on the geoparsing task, defined as identifying toponyms within textual content and subsequently disambiguating them, e.g., obtaining corresponding GPS coordinates. They compare various models across established assessment datasets tailored for evaluating this particular task, such as GeoCorpora (Wallgrün et al., 2018), GeoVirus (Gritta et al., 2018) or WikToR (Gritta et al., 2018). Their findings suggest that models generally produce accurate predictions in Western countries (such as the U.S. and the UK), while poorer prediction quality tends to occur predominantly outside those countries.

An alternative downstream task worth considering is text classification. A particularly interesting dataset is proposed by Humaid (Alam et al., 2021), which aims to categorize tweets according to relevance for crisis managers during disaster situations. The labels used for the classification include displacement and evacuation, infrastructure and utility damage, and rescue volunteerism or charitable efforts. Researchers have collected and manually annotated more than 77,000 tweets of 19 distinct natural disasters that occurred between 2016 and 2019 in multiple countries. They collected only tweets in English that were located in the affected areas, using tweet geo-coordinates when available; otherwise, user location was used. The authors trained several models on each of these events, including four language models, to classify and tag the tweets into the labels. Table 2 shows the scores of these models depending on the events. We add a final column containing the average score of the language models. Events are ranked from the highest classification score to the lowest. The location of the disaster in underrepresented countries does not seem to affect the performance of the classifiers, as evidenced by the fact that disasters ranked second or third occurred in Ecuador and Mexico, countries that are in world regions with fewer representations and higher semantic isolation, as depicted in Fig. 2. The nature of the event appears to play a much more significant role in determining the scores. Notably, events related to hurricanes tend to exhibit the lowest scores, despite occurring in the United States. Nevertheless, the granularity of geographical entities present in the tweets gathered by Humaid varies depending on the locations of the disasters. As indicated by (Suwaileh et al., 2023), disasters occurring outside the U.S. or Europe refer to locations on a larger scale, typically at the national level (approximately 85% for the Ecuador earthquake and 70% for the Mexico earthquake refer only to Ecuador and Mexico, respectively), in contrast to hurricanes in the U.S., for which approximately 15% refer to the U.S. Indeed, disasters taking place in Western countries showcase a wider variety of locations. Tweets associated with these events contain street names, states, and descriptions of natural landscapes, which could impact classifications.

**Table 2** Ranking of events based on the average of classifier scores as computed and presented by (Alam et al., 2021)

| Event | BERT | D-BERT | RoBERTa | XLM-RoBERTa | Mean Models |
|---|---|---|---|---|---|
| 2016 Italy Earthquake | 0.871 | 0.878 | 0.885 | 0.877 | 0.878 |
| 2016 Ecuador Earthquake | 0.861 | 0.872 | 0.872 | 0.866 | 0.868 |
| 2017 Mexico Earthquake | 0.845 | 0.854 | 0.863 | 0.847 | 0.852 |
| 2019 Pakistan Earthquake | 0.820 | 0.822 | 0.834 | 0.827 | 0.826 |
| 2016 Hurricane Matthew | 0.786 | 0.780 | 0.815 | 0.784 | 0.791 |
| 2019 Cyclone Idai | 0.790 | 0.779 | 0.796 | 0.793 | 0.789 |
| 2016 Canada Wildfires | 0.792 | 0.781 | 0.791 | 0.768 | 0.783 |
| 2018 Greece Wildfires | 0.788 | 0.739 | 0.783 | 0.783 | 0.773 |
| 2018 Hurricane Florence | 0.768 | 0.773 | 0.780 | 0.765 | 0.771 |
| 2018 California Wildfires | 0.760 | 0.767 | 0.764 | 0.757 | 0.762 |
| 2016 Kaikoura Earthquake | 0.768 | 0.743 | 0.765 | 0.760 | 0.759 |
| 2017 Hurricane Harvey | 0.759 | 0.743 | 0.763 | 0.761 | 0.756 |
| 2017 Sri Lanka Floods | 0.703 | 0.763 | 0.727 | 0.798 | 0.748 |
| 2018 Maryland Floods | 0.697 | 0.734 | 0.760 | 0.798 | 0.747 |
| 2018 Kerala Floods | 0.732 | 0.732 | 0.745 | 0.746 | 0.739 |
| 2019 Midwestern Floods | 0.702 | 0.706 | 0.764 | 0.726 | 0.724 |
| 2017 Hurricane Irma | 0.722 | 0.723 | 0.730 | 0.717 | 0.723 |
| 2017 Hurricane Maria | 0.715 | 0.722 | 0.727 | 0.723 | 0.722 |
| 2019 Hurricane Dorian | 0.691 | 0.691 | 0.686 | 0.691 | 0.690 |

To summarize, very few datasets enable us to assess geographical knowledge disparities. Moreover, even when these datasets can easily be segmented based on location, the granularity of toponyms turns out to be biased as well, unfortunately. Granularity diversity across locations is not uniformly distributed worldwide. Regions where geographical knowledge in models is most limited coincide with regions where we find poverty in fine-grained location granularity within traditional NLP benchmarks.

## 5 Discussion

In this section, we present some challenges related to the inequalities in the quality of geographic knowledge of models for humanitarian action. Afterward, we list various classic methods for improving CLMs and their possible adaptations to our specific theme.

### 5.1 Challenges

Overall, CLMs have very good representations of various aspects of geography. Unfortunately, as illustrated by our four indicators, the quality of knowledge is not evenly and equitably distributed globally. CLMs possess very precise knowledge of Western Europe and Oceania. This is also true, albeit to a lesser extent, for North America, South America, and Asia. With respect to Africa, Eastern Europe, Southeast Asia, and island countries,

CLMs generally have more random knowledge, leading to undoubtedly higher rates of hallucinations.

While the massive use of CLMs raises many concerns, such as the degradation of the job market for highly qualified or artistic professions (Zarifhonarvar, 2024), the negative impact on student learning (Gajos & Mamykina, 2022), or in our field, the congestion of the peer-reviewing system (Kobak et al., 2024; Liu & Brown, 2023), CLMs can contribute to tasks for social good, such as humanitarian efforts, by facilitating information gathering for crisis management (Palen & Anderson, 2016). Unfortunately, the countries most affected by climate change, with respect to vulnerability and fatalities, are also countries where the CLMs evaluated in this study have the poorest geographic knowledge. Successive droughts, extreme flood episodes, hurricanes & cyclones, and rising sea levels particularly impact tropical countries (Donatti et al., 2024). Additionally, CLMs are often trained on large corpora in English, with other languages being marginally represented, as illustrated in Table 1. However, the most vulnerable countries subject to hazards are native English-speaking ones.

Moreover, as discussed in the Sect. 4.2 Implications for Downstream Applications, assessing the impact of limited geographic knowledge in countries that are likely to be underrepresented in training datasets is challenging. Indeed, the datasets collected in these areas often contain very general mentions of locations (at the country or city level) but rarely provide finer levels of description, such as neighbourhoods or points of interest.

### 5.2 Strategies to overcome these challenges

How can we improve the geographic knowledge of models so that their use in humanitarian aid tasks yields better results, especially for countries underrepresented in training? First, it is essential to create a benchmark dataset containing locations that are evenly distributed across the same levels in the spatial hierarchy (point of interest/neighbourhood/city/state/country) and with a significant proportion of fine granularity. This dataset should include both Western countries and countries that are currently underrepresented in the data. It should also be labelled for various tasks, such as toponym (proper name of a location) identification with geocoding (assigning GPS coordinates to locations), text classification, event detection, and event deduplication.

Indicators 3 and 4 (which focus on biases in the representation of distances) reveal that cities in underrepresented regions may rarely co-occur in training datasets. According to Louwerse and Zwaan (2009), it is possible to estimate the geographic distance between two cities on the basis of their co-occurrence in large text corpora. Therefore, our hypothesis is that since the models fail to accurately reflect geographic proximity in their embeddings for underrepresented areas, cities in these regions rarely co-occur with each other in the data. Furthermore, we hypothesize that underrepresented cities are more likely to co-occur with foreign cities, i.e., cities from other countries or even other continents, as illustrated in Figs. 9 and 10 in Appendix A. To improve the correlation between geographic and semantic distances, training datasets should be enriched with co-occurrences of geographically proximate cities. One possible approach would be to generate synthetic data using sources such as OpenStreetMap. However, this method has limitations, as OpenStreetMap itself exhibits geographic bias, since its geographic entities are less complete for underrepresented countries. With respect to masked language models (MLMs), it may be feasible to retrain the models by masking geographically informative words when they co-occur with other place names. Nevertheless, care must be taken to ensure that this form of geographically oriented fine-tuning does not degrade the model's performance on other tasks. As shown by the detailed results of the third indicator (Appendix A.3), the GeoLM model, fine-tuned on sentences generated from OpenStreetMap data, captures geographic distances much more effectively. However, it has lost much of its language understanding capabilities, yielding near-zero performance on the second indicator (Appendix A.2).

For CLMs, the most straightforward and least computationally intensive approach is prompt engineering (White et al., 2023). In this context, its goal is to enrich prompts with static geographic information, meaning that the same information is used for each prompt. This method can be supplemented by few-shot learning, which involves providing a few examples for the task in question along with their answers (Brown et al., 2020). Instead of using static geographic information, retrieval-augmented generation (RAG) can be employed (Balaguer et al., 2024). This approach involves splitting the dataset into training/testing sets, even though it is not strictly a machine learning training process. The training dataset is vectorized and saved in an embedding database. For each test element, the RAG mechanism aims to enrich the prompt with the n closest phrases from the training set to provide precise and contextual examples to the language model via its prompt.

The third approach comes from agentic CLM techniques, which allow language models to interact with external resources if they are not confident enough with their predictions (Xi et al., 2023). In our context of unequal-quality geographic knowledge across countries, the goal is to enable language models to query gazetteers (such as GeoNames[7] or OpenStreetMap[8]) to retrieve data describing geographical objects they are not familiar with.

The fourth approach consists of fine-tuning a language model on corpora rich in geographical information, such as Wikipedia, which, as evidenced by BERT's performance in this study, appears to facilitate the acquisition of geographic knowledge. Another possibility for reducing inequalities in the distribution of geographic knowledge is to fine-tune models on multilingual datasets, illustrated in this paper with m-BERT, which includes underresourced languages. Several research groups are already proposing approaches or datasets on underrepresented languages, such as for Southeast Asia (Lovenia et al., 2024) and Africa (Adebara & Abdul-Mageed, 2022). Moreover, it would be interesting to assess the contribution of local languages in multilingual models (Li et al., 2024) to evaluate the enrichment of geographical knowledge brought by these languages.

The fifth method, which requires minimal resources (mainly RAM), is model merging. This involves combining, layer by layer, two or more language models. Various merging methods exist (Yadav et al., 2023; Yu et al., 2024), offering different strategies for selecting certain weights, averaging them, or concatenating them. Thus, we can consider merging models with good geographic knowledge with models that are highly performant in text generation and question answering.

The final method is called mixture of experts (MoE) (Jiang et al., 2024). Multiple models collaborate to answer users' questions, with each model specializing in a specific topic or task. For each question, the MoE selects one or more outputs from expert models (sparsity).

[7] https://www.geonames.org

[8] https://www.openstreetmap.org

It is indeed possible to combine several or all of these approaches. However, the cost of these operations must be weighed against the improvement obtained.

## 6 Conclusion

This study examined geographic biases in state-of-the-art language models, revealing significant disparities in their ability to represent spatial knowledge across different regions of the world. Our analysis shows that current models are largely centred on Western countries, reflecting an underlying geographic imbalance in training data that constrains their applicability and fairness for diverse populations. These results highlight the importance of actively addressing geographic underrepresentation in pretraining corpora.

While our investigation focused primarily on English-language data because of its predominance in available datasets, we acknowledge that geographic bias is a complex issue that transcends linguistic boundaries. Future research should therefore extend evaluations to multilingual models and develop standardized benchmark datasets that more comprehensively capture geographic diversity. Such efforts are critical for understanding and mitigating the nuanced effects of spatial bias in natural language processing applications. In addition, future studies should assess the reasoning capabilities of causal language models to determine whether geographic biases persist in more complex tasks.

Overall, our work highlights a previously underexplored dimension of bias in language models and represents a crucial step towards building more inclusive models that account for geographic context. To ensure safety and reliability in applications, particularly in the context of increasingly frequent global crises, language models must perform equally across all regions of the world.

## Detailed experiments and results

This section details the results obtained from the four indicators previously mentioned for the ten widely used models outlined in Table 1.

### Indicator 1: spatial information coverage in training datasets

#### Key results

MLMs: The multilingual model m-BERT has broader global geographic coverage, with a slight reduction in coverage for Anglophone countries. However, it is highly sensitive to prompt formulation. XLM-RoBERTa, on the other hand, does not demonstrate improved geographic coverage. CLMs: This indicator is not effective for evaluating the geographic coverage of their training data.

### Detailed results

This indicator proposes an indirect evaluation of the training corpus by counting the number of cities with populations exceeding 100,000 that appear in the models' fixed vocabularies. These vocabularies are constructed from the most frequently encountered words during training. The results for all the models are provided in Table 3. Figures 3 and 4 present the percentage of cities with more than 100,000 inhabitants, either aggregated by country (Fig. 3) or by 5°×5° spatial grid cells (Fig. 4). Note that we used the Ada model instead of ChatGPT (GPT−3.5) because its embeddings are not accessible.

The first interesting observation concerns the MLM. BERT and m-BERT have the most cities in terms of their vocabularies. Another interesting result is that m-BERT's training allowed it to find more cities outside the English-speaking world, which led to a loss of coverage for Oceania and North America. m-BERT performs better on continents where English is not a native language because of its more diverse training corpus. An analysis of the results reveals that m-BERT is highly sensitive to the prompt. Even minor modifications can cause significant fluctuations in its performance. When capital prompts were applied to cities with more than 100,000 inhabitants, m-BERT remained competitive with BERT. The fine-tuning of m-BERT likely impaired its ability to respond accurately to questions in English (Choshen et al., 2022), which may explain its decrease in performance for cities with more than 100,000 inhabitants. Very few city names are included in the models' fixed vocabularies, except in Oceania, Europe, and North America, even among multilingual models.

However, in the case of CLMs, analysing the spatial coverage of training data solely on the basis of vocabulary content is not feasible. There are very few cities in their vocabularies. Moreover, according to manual analysis, the cities that appear in the most frequent words of the models are cities that are spelled like common words, such as Bath (United Kingdom). LLaMa and Mistral use SentencePiece (Kudo & Richardson, 2018) with byte pair encoding algorithms for their tokenizer, which means that they can handle different alphabets (Latin, Chinese, Greek, etc.). Thus, their fixed vocabulary is composed mainly of subtokens, which are frequently encountered syllables from the training process. Therefore, this indicator is not useful when working with CLMs. To our knowledge, there is currently no alternative method for assessing the extent of geographical information present in training corpora when those corpora are not publicly available. This remains an open challenge.

Mistral has been partially trained with CommonCrawl, a publicly available dataset of web page data collected by web crawling. Ilyankou et al. (2024), observing the good performance of CLMs on geographical tasks, attempted to evaluate the percentage of CommonCrawl documents containing postal addresses or GPS coordinates. In their sample, 18% of the documents contain at least one of these two types of information, with the majority being addresses or GPS coordinates within Google Maps links. Unfortunately, it is currently impossible to know how many epochs the models were trained on CommonCrawl and what other training datasets were used. We also do not know whether CommonCrawl is the only source of geographical information for training.

(a) BERT (b) m-BERT (c) GeoLM

(d) RoBERTa (e) XLM-RoBERTa (f) Mistral

(g) Mistral v0.3 (h) LLaMA 3 (i) LLaMA 2

(j) ChatGPT (k) Legend

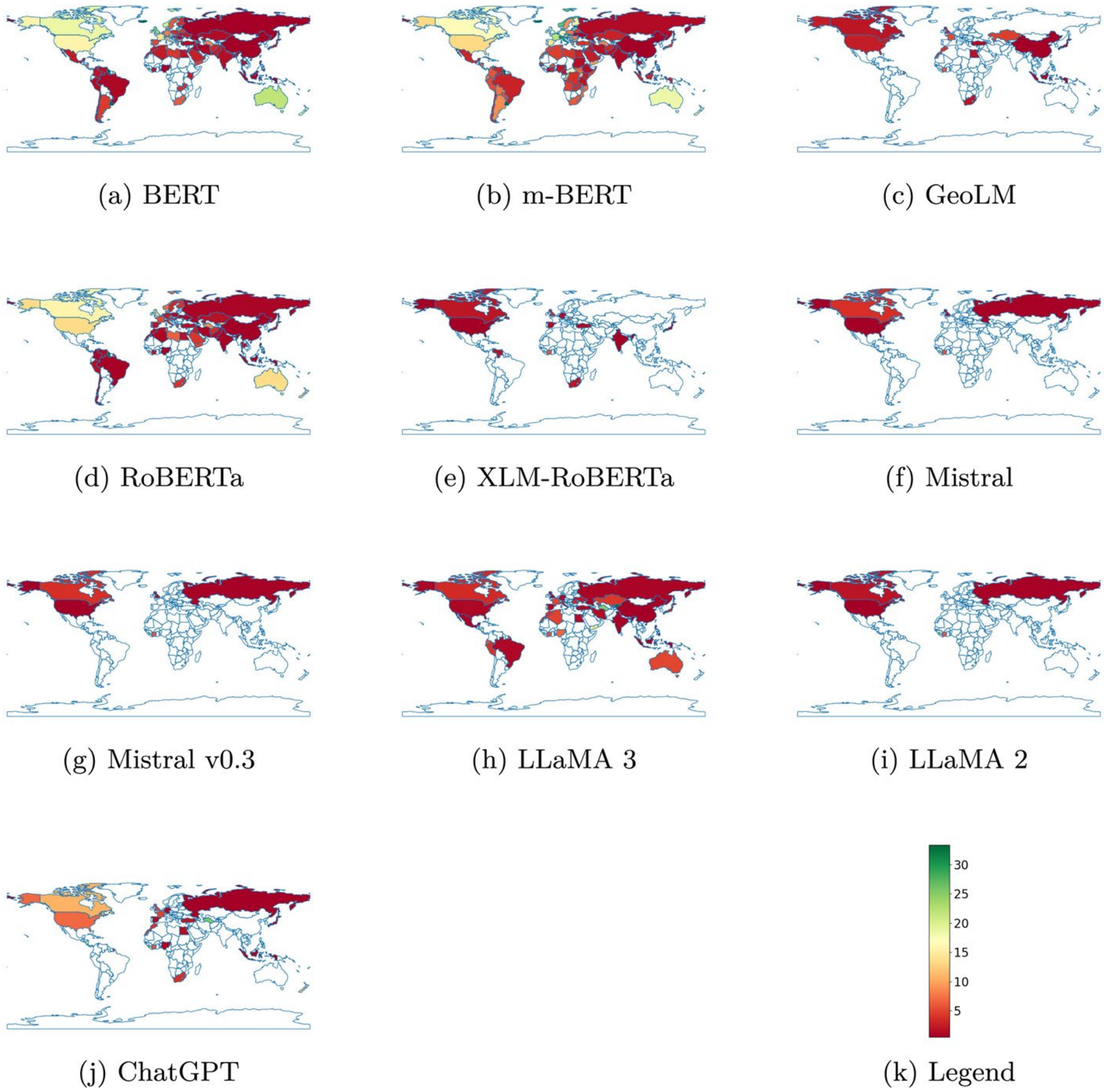

**Fig. 3** Percentages by countries of cities with more than 100K inhabitants in the models' vocabulary. Countries in white have none of their cities in the vocabulary

## Indicator 2: spatial disparities in geographical knowledge

### Key results

Language models perform poorest on Africa and show limited accuracy for Eastern Europe, Central Asia and Southeast Asia. Knowledge about island nations is particularly scarce. This indicator has one limitation: it is affected by cities with homonyms in other countries.

### Detailed results

To evaluate geographical knowledge, the models are compared by predicting the country from its cities with more than 100,000 inhabitants. The correct answers aggregated by country are shown in Figs. 5 and 6 by pixels of 5°×5°, i.e., each pixel shows the percentage of

(a) BERT

(b) m-BERT

(c) GeoLM

(d) RoBERTa

(e) XLM-RoBERTa

(f) Mistral

(g) Mistral v0.3

(h) LLaMA 3

(i) LLaMA 2

(j) ChatGPT

(k) Legend

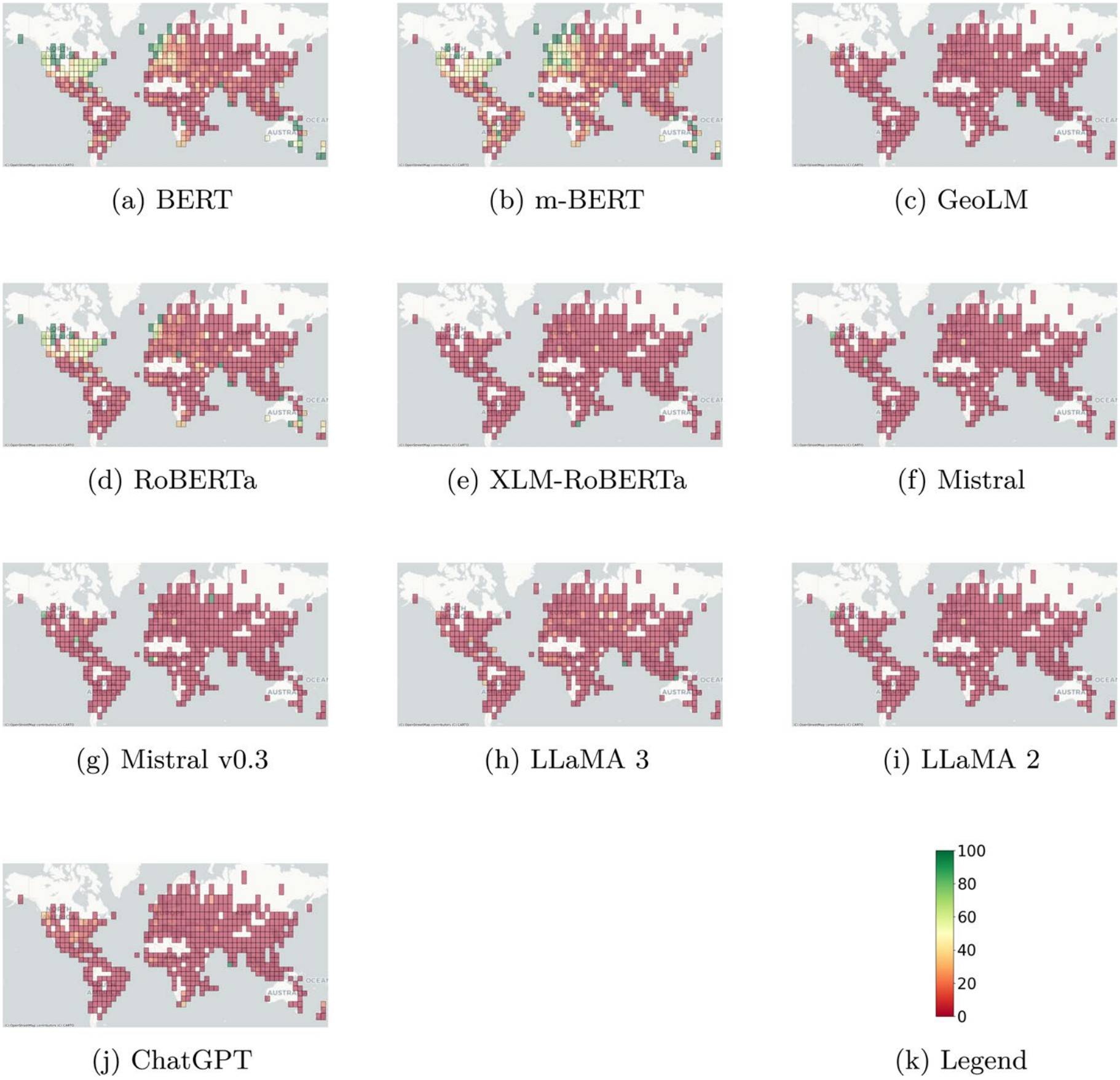

**Fig. 4** Detailed results showing the percentages of cities in the vocabulary by spatial aggregation in pixels of 5°×5°

**Table 3** Percentages of cities with more than 100K inhabitants in the models' vocabulary, averaged by continent

| | N.Am | S.Am | Eur. | Africa | Asia | Oceania |
|---|---|---|---|---|---|---|
| BERT | **33.80** | 4.82 | 22.88 | 4.98 | 4.50 | **60.00** |
| m-BERT | 30.05 | **13.00** | **36.05** | **10.82** | **6.16** | 50.00 |
| GeoLM | 1.41 | 0.00 | 0.67 | 0.69 | 0.26 | 0.00 |
| RoBERTa | 26.92 | 1.89 | 9.38 | 3.26 | 2.75 | 36.67 |
| XLM-RoBERTa | 0.31 | 0.21 | 0.67 | 0.69 | 0.17 | 0.00 |
| LLaMa 2 | 0.31 | 0.00 | 0.22 | 0.34 | 0.00 | 0.00 |
| LLaMa 3 | 1.41 | **1.05** | **1.79** | **2.23** | **1.00** | **3.33** |
| Mistral-v01 | 0.47 | 0.00 | 0.33 | 0.34 | 0.00 | 0.00 |
| Mistral-v03 | 0.47 | 0.00 | 0.33 | 0.34 | 0.00 | 0.00 |
| ChatGPT | **4.85** | 0.00 | 1.34 | 1.20 | 0.39 | **3.33** |
| Nb of cities | 639 | 477 | 896 | 582 | 2289 | 30 |

*N. Am* North America, *S. Am* South America, and *Eur.* Europe

The best results by model family (MLMs and CLMs) are in bold accurate predictions by averaging the output for each city located in the pixel. Table 4 provides the results in percentages aggregated by continent.

The first observation is that the quality of geographical knowledge is not directly related to the number of parameters in the models. For instance, BERT (110 million) achieves more accurate predictions than Mistral (7 billion). The training data appear to play a crucial role in this ability. We hypothesize that BERT, trained on 2.5 billion words from Wikipedia and 800 million from Google Books, acquired this type of geographical knowledge during its training. Mistral may have been less exposed to geographical information during its training. Although its architecture gives it greater memorization capacity, it remains less effective than BERT (with a difference of -7 points with v01 and -3 points with v03, as shown in column *World* in Table 4). This suggests that the bias introduced by the training dataset is among the biases that have the greatest impact on downstream performance (Navigli et al., 2023).

While the poor results in North America and Oceania can be related to the fact that both contain island countries, the indicator values are lower for Africa across all the models except for ChatGPT, as illustrated in Fig. 5.

Another interesting observation is that multilingual models, despite capturing a greater diversity of languages, do not necessarily improve upon the base models. XLM-RoBERTa even performs worse than RoBERTa. However, multilingual BERT predictions are better for the European and African continents, albeit at the expense of Asia. Finally, the results of GeoLM are 0. We contacted the authors,[9] but we did not receive a response. The predictions for the masked country are words without geographical information, such as *archived* or *Metacritic*. These two words alone account for more than 40% of the predictions of GeoLM.

Now, let us consider more cities by asking the models to guess the country of cities with a population of more than 100,000 inhabitants. This experiment enables us to obtain a finer spatial granularity of the spatial distribution of the geographical knowledge quality of the models. The results are aggregated by continent in Table 5 and by pixels of $5° \times 5°$ in Fig. 7.

Compared with the capital test, the first observation is that the scores of encoder-type models decrease, with m-BERT losing more than 30 points (Fig. 5, column *World*), while the least performant models on capital cities (XLM-RoBERTa and RoBERTa) have the smallest decrease in score. RoBERTa even outperforms m-BERT. The scores of CLMs do not all deteriorate. Even though LLaMa loses 6–7 points and ChatGPT loses 2 points, Mistral models improve their performance by up to +15 points for v0.3, with North America, Oceania, and Asia observing the best progression. One possible explanation is that the Mistral models' poor knowledge of insular countries on these continents has weighed down their score on the capital city test.

Another observation is that Africa remains the least well-understood continent by language models, as shown in Table 5. On the maps in Fig. 7, we can also see that the models have less accurate knowledge for Eastern Europe, Central Asia, Southeast Asia, and the North American continent.

Surprisingly, the models fail to correctly link North American cities to their country, given the intuition that they were trained on corpora from this geographical area. After an analysis in Appendix B for North America, it appears that many of these cities have homonyms, especially in Europe, such as Paris, Rome or Athens, as shown in Fig. 11. According to Zelinsky (1967), this can be explained by the desire of European colonists to transpose

[9] https://huggingface.co/zekun-li/geolm-base-cased/discussions/2

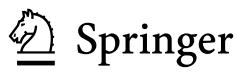

(a) BERT

(b) m-BERT

(c) GeoLM

(d) RoBERTa

(e) XLM-RoBERTa

(f) Mistral

(g) Mistral v0.3

(h) LLaMA 3

(i) LLaMA 2

(j) ChatGPT

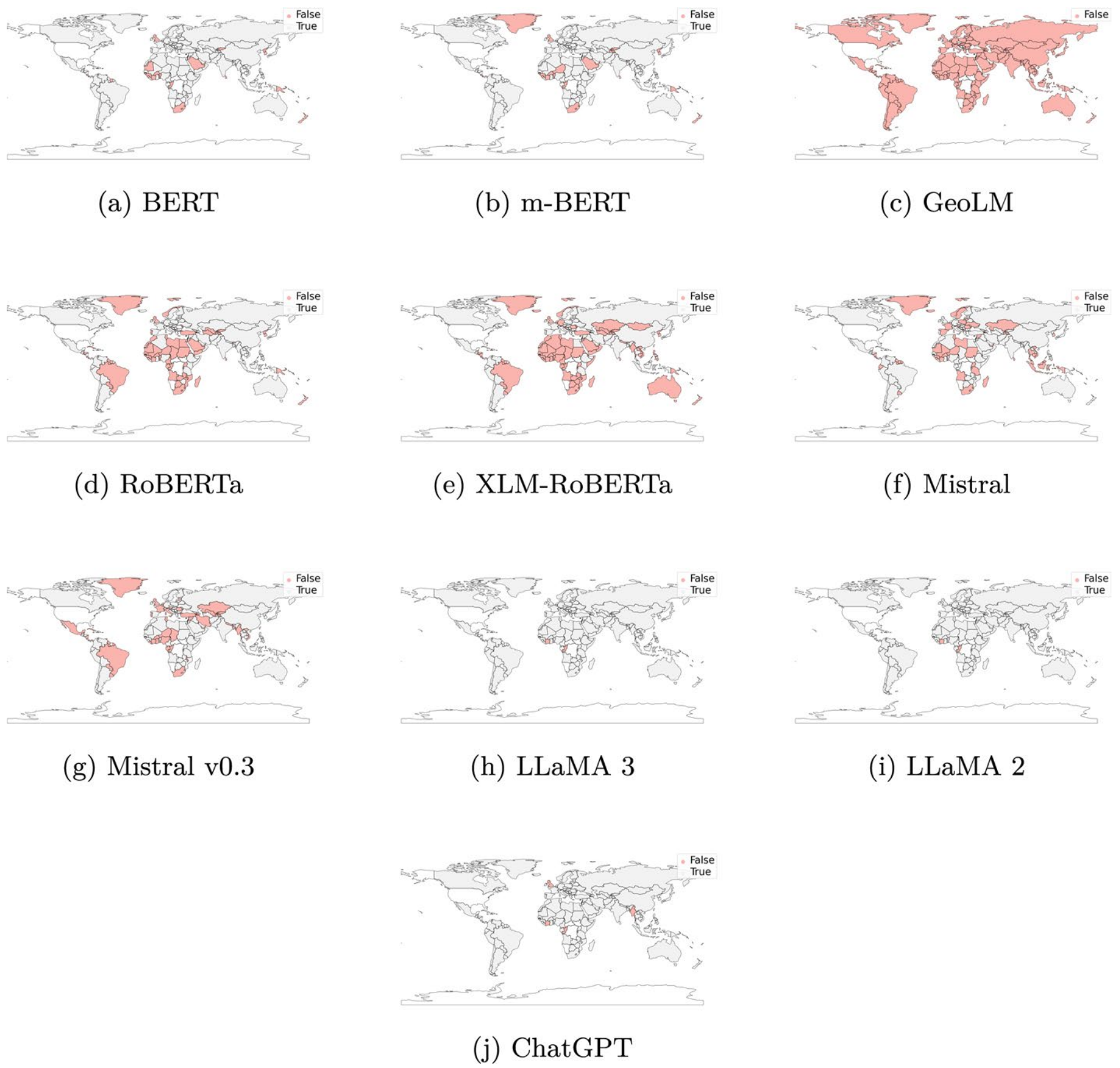

**Fig. 5** Correct predictions of the country given its capital

their cultural heritage onto the new continent in the 19th century. Thus, the models tend to attribute these ambiguous American toponyms to their original countries in Europe when no other context is provided. Therefore, if these cities were accompanied by other toponyms, the models would achieve better scores (Kafando et al., 2023).

### Indicator 3: correlation between geographic distance and semantic distance

### Key results

There is little correlation between geographic and semantic distances, except for OpenAI's embedding model and the GeoLM MLM retrained on Wikipedia-based sentences. However, stronger correlations are observed for Europe and Asia.

(a) BERT (b) m-BERT (c) GeoLM

(d) RoBERTa (e) XLM-RoBERTa (f) Mistral

(g) LLaMA 2 (h) LLaMA 3 (i) Mistral v0.3

(j) ChatGPT (k) Legend

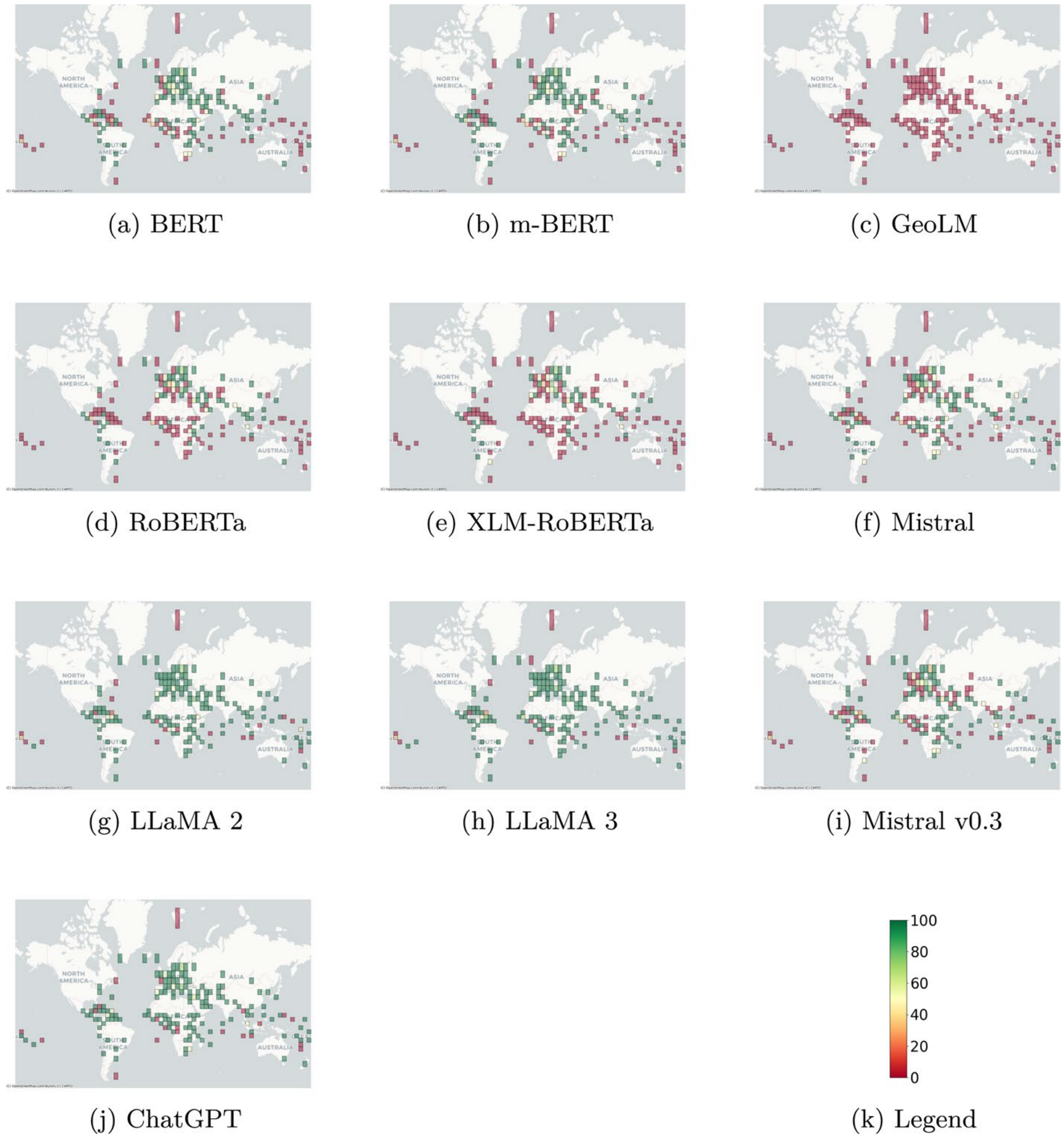

**Fig. 6** Correct capital predicted and location

## Detailed results

We compare the correlation between semantic and geographical distances pairwise for the 50 most populated cities by continent. This type of result for the BERT model for Europe is shown in Fig. 8. We apply a linear regression between geographic and semantic distances and display the $R^2$ of the linear regression for all the models in Table 6. Note that we used the Ada model instead of ChatGPT (GPT$-$3.5) because we could not access its embeddings.

First, the correlation coefficients ($\in [0, 1]$) are low for all the models. This finding indicates that the models' representations (or embeddings) struggle to accurately capture the geographical distances between locations, as suggested by Decoupes et al. (2023). However, for MLMs, GeoLM is competitive with the best CLM in this task. GeoLM's specific training from sentences generated from OpenStreetMap, containing neighbours of locations, allows it to grasp geographical distances better. Once again, the training dataset is crucial.

**Table 4** Percentages of correct predictions at the continental and global levels

| | N.Am | S.Am | Eur. | Africa | Asia | Ocea. | World |
|---|---|---|---|---|---|---|---|
| BERT | **35.14** | 78.57 | 74.0 | 55.56 | **75.47** | **16.67** | **57.94** |
| m-BERT | 32.43 | **85.71** | **78.0** | **61.11** | 66.04 | **16.67** | **57.94** |
| GeoLM | 0.00 | 0.00 | 0.0 | 0.00 | 0.00 | 0.00 | 0.00 |
| RoBERTa | 16.22 | 57.14 | 60.0 | 18.52 | 54.72 | 4.17 | 36.05 |
| XLM-RoBERTa | 13.51 | 50.00 | 50.0 | 7.41 | 35.85 | 0.00 | 25.75 |
| LLaMa 2 | 64.86 | **92.86** | **94.0** | 83.33 | **84.91** | 62.50 | 81.12 |
| LLaMa 3 | **72.97** | **92.86** | 90.0 | 88.89 | 88.68 | **79.17** | **85.41** |
| Mistral-v01 | 40.54 | 57.14 | 54.0 | 55.56 | 60.38 | 29.17 | 51.07 |
| Mistral-v03 | 40.54 | 57.14 | 60.0 | 66.67 | 52.83 | 45.83 | 54.94 |
| ChatGPT | **75.68** | 85.71 | 82.0 | **90.74** | 83.02 | **75.00** | **82.40** |
| Nb countries | 37 | 14 | 50 | 54 | 53 | 24 | 232 |

*N. Am* North America, *S. Am* South America, *Eur.* Europe, and *Ocea.* Oceania

For each continent, the best results by model family (MLMs and CLMs) are in bold

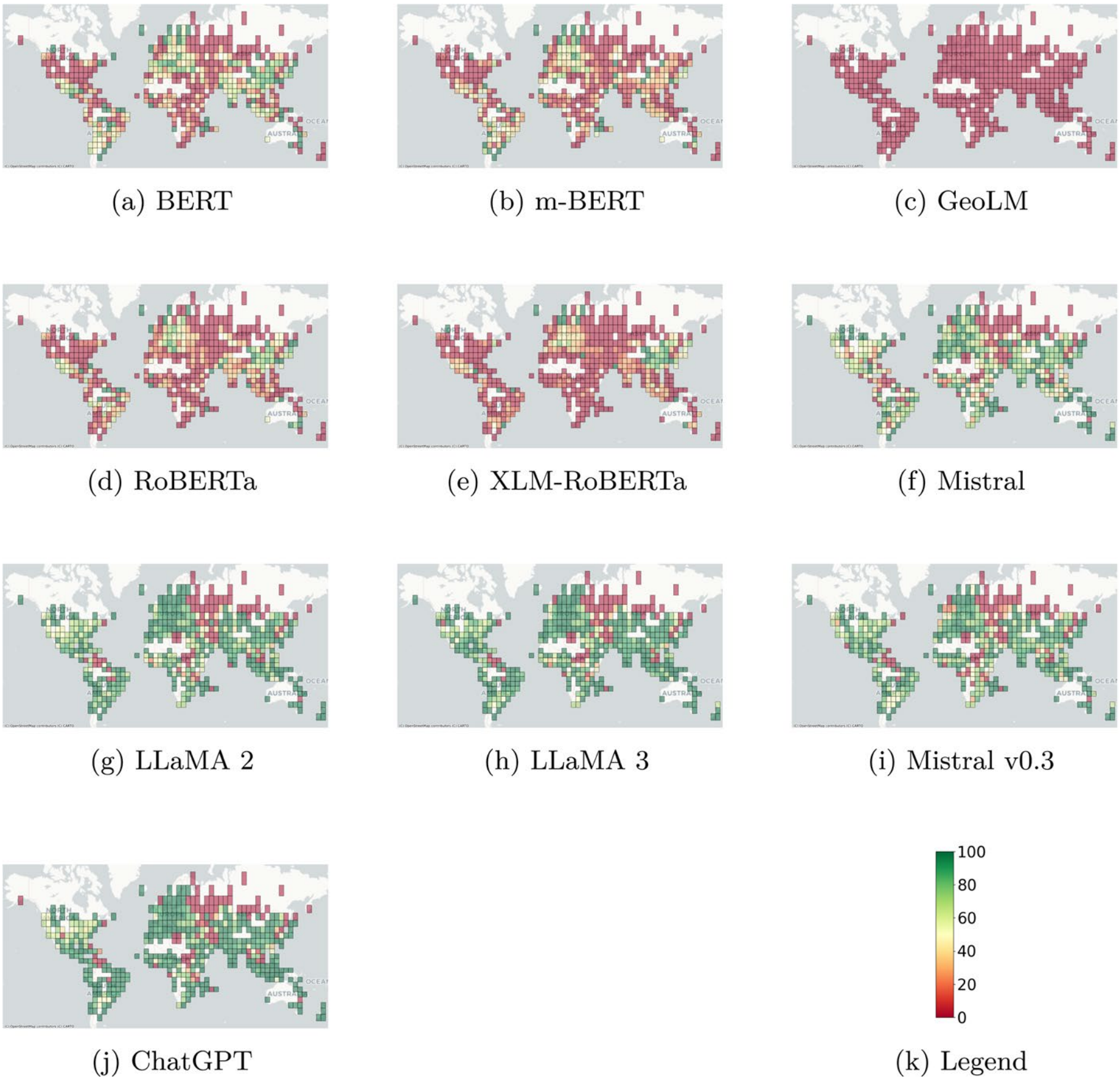

(a) BERT

(b) m-BERT

(c) GeoLM

(d) RoBERTa

(e) XLM-RoBERTa

(f) Mistral

(g) LLaMA 2

(h) LLaMA 3

(i) Mistral v0.3

(j) ChatGPT

100
80
60
40
20
0

(k) Legend

**Fig. 7** Percentages of correct country predictions given cities with more than 100K inhabitants by spatial aggregation in pixels of $5° \times 5°$

**Table 5** Percentages of correct country predictions based on city names with more than 100K inhabitants

| | N.Am | S.Am | Eur. | Africa | Asia | Ocea. | World |
|---|---|---|---|---|---|---|---|
| BERT | **21.28** | 32.91 | **37.17** | **19.24** | **52.91** | 33.33 | **39.87** |
| m-BERT | 15.65 | **35.64** | 32.37 | 18.73 | 25.03 | 20.00 | 25.41 |
| GeoLM | 0.00 | 0.00 | 0.0 | 0.00 | 0.00 | 0.00 | 0.00 |
| RoBERTa | 16.28 | 24.74 | 32.48 | 12.37 | 37.09 | **36.67** | 29.39 |
| XLM-RoBERTa | 9.39 | 13.00 | 24.89 | 4.12 | 34.21 | 16.67 | 23.54 |
| LLaMa 2 | 72.46 | 71.91 | **70.98** | 68.56 | 78.94 | **96.67** | 74.82 |
| Llama3 | **78.4** | 76.52 | 70.54 | 70.62 | 81.39 | 93.33 | 77.32 |
| Mistral-v01 | 52.11 | 53.88 | 58.15 | 55.33 | 65.71 | 86.67 | 60.29 |
| Mistral-v03 | 73.87 | 68.34 | 60.94 | 66.15 | 75.32 | 90.0 | 70.81 |
| ChatGPT | 75.12 | **80.08** | 70.87 | **80.58** | **84.32** | 93.33 | **79.84** |
| Nb of cities | 639 | 477 | 896 | 582 | 2289 | 30 | 4913 |

*N. Am* North America, *S. Am* South America, and *Eur.* Europe

The best results by model family (MLM and CLMs) are in bold

With respect to CLMs, LLaMa 2 did not capture geographical distances, whereas Ada and OpenAI's embedding model achieved the highest scores. The final observation is that Europe, Asia, and, to a lesser extent, North America are the continents whose geographical distances are best captured. However, these continents, in the arrangement of their countries, do not share the same characteristics. Indeed, Europe is the smallest continent with the most small countries, while North America is a continent that contains large countries and island countries (from Central America). Apart from the Dominican Republic, Haiti, and Jamaica, it is rare for island states to have cities with more than 100,000 inhabitants. Indeed, there are only 33 cities with more than 100,000 inhabitants in such states, representing 33 out of 639 or 5.16% of the cities with more than 100,000 inhabitants on this continent. Therefore, these island states are underrepresented in our analysis, leaving room for the giants of the continent: Canada, the United States, and Mexico.

Thus, the characteristics of continents do not seem to explain the observed differences between them. With respect to the results from the first indicator, a correlation between semantic and geographic distances could be influenced by the over- or underrepresented areas. First, when a location is overrepresented in the training data, it leads to a smaller semantic distance to other locations. Second, the opposite occurs: when a location is underrepresented in the training data, its embeddings are more distant from the embeddings of other locations. Indicator 4, as detailed below, determines whether countries are distant or in the centre of the semantic space.

## Indicator 4: anomaly between geographical distance and semantic distance

### Key results

Geographic distances in semantic space are distorted: Western countries appear unnaturally close, especially for Oceania cities, while African cities are abnormally far apart. In practice, African cities are semantically closer to Western cities than to their neighbouring cities.

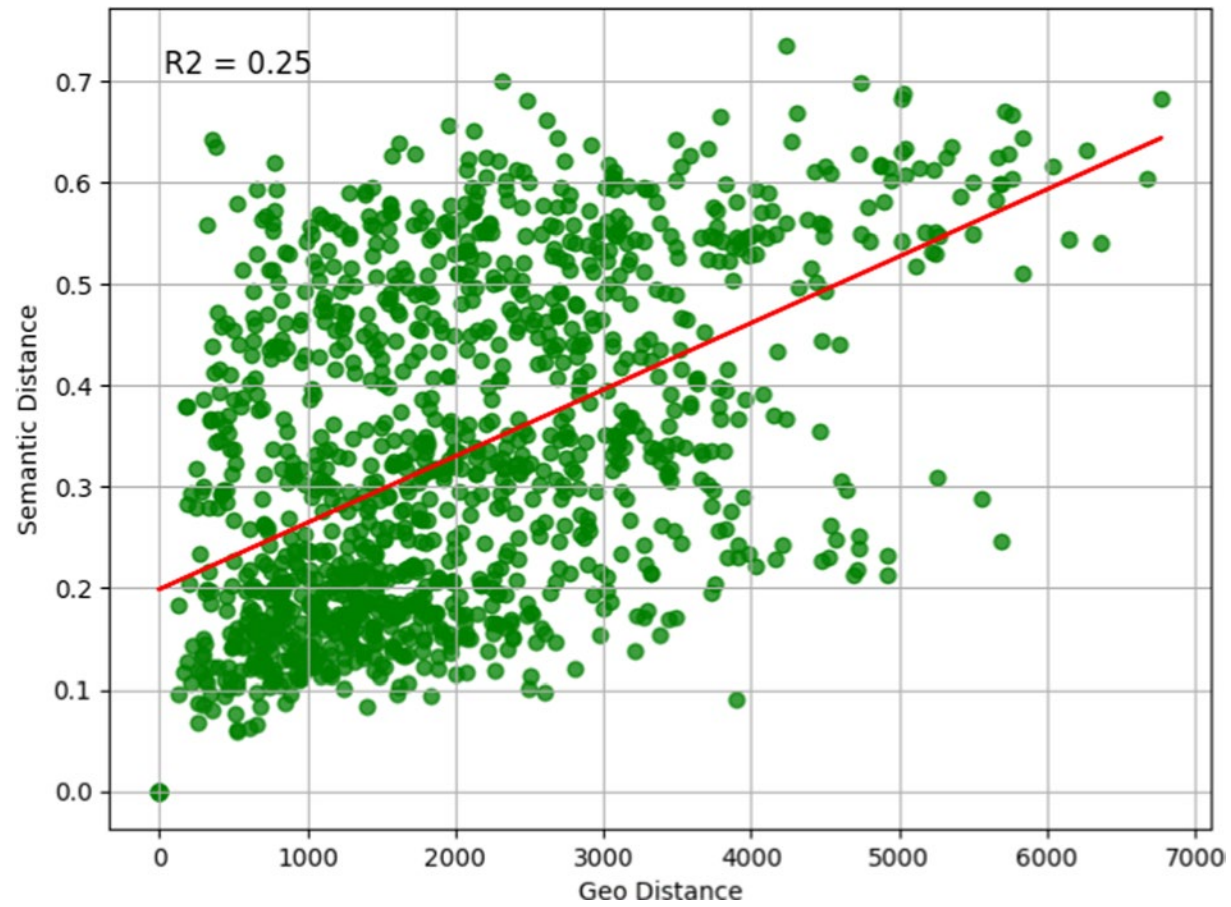

**Fig. 8** Linear regression between geographical distance (km) and semantic distance with BERT for Europe

**Table 6** $R^2$ of the linear regression between geographical distance and semantic distance between pairs of cities

| | N.Am. | S.Am. | Europe | Africa | Asia | Oceania |
|---|---|---|---|---|---|---|
| BERT | 0.08 | **0.07** | 0.25 | 0.07 | 0.09 | 0.03 |
| m-BERT | 0.01 | 0.01 | 0.14 | 0.03 | 0.02 | 0.00 |
| GeoLM | **0.17** | 0.03 | **0.37** | **0.09** | **0.32** | **0.07** |
| RoBERTa | 0.03 | 0.02 | 0.13 | 0.02 | 0.05 | 0.00 |
| XLM-RoBERTa | 0.05 | 0.00 | 0.19 | 0.01 | 0.00 | 0.00 |
| LLaMa 2 | 0.07 | 0.06 | 0.15 | 0.06 | 0.15 | 0.04 |
| LLaMa 3 | 0.08 | 0.05 | 0.18 | 0.08 | 0.27 | 0.10 |
| Mistral-v01 | 0.07 | 0.07 | 0.20 | 0.07 | 0.15 | 0.05 |
| Mistral-v03 | 0.09 | 0.07 | 0.22 | 0.08 | 0.16 | 0.06 |
| Ada | **0.17** | **0.20** | **0.28** | **0.17** | **0.38** | **0.17** |

*N. Am* North America and *S. Am* South America

The best results by model family (MLMs and CLMs) are in bold

## Detailed results

As shown by indicator 3, some continents, such as Africa and Oceania, have very low correlations between semantic and geographical distances. With this new and final indicator, we propose to explain these low correlations either by an overrepresentation of countries in the training datasets, positioning the places of these countries in the centre of the semantic space, or, on the contrary, by an underrepresentation of these places, isolating them in this space.

Figure 2e shows, for the BERT model, the country average of the semantic distances for its three most populous cities with the most populous cities in the world. In Africa, we find countries that are the most semantically distant, such as Burkina Faso, the Democratic Republic of the Congo and Mauritania. On the other hand, the countries of Oceania, such as Australia and New Zealand, are abnormally close semantically to the most populous cities in the world compared with their geographical distance.

To validate these observations for all the models, we use the GDI ratio (semantic distance divided by geographic distance). Table 7 shows, by model and continent, the average GDI and the number of countries in the 20 most or least distant countries. The first observation is that the trends described below are consistent across the models. For all the models, Europe is the most distant continent from the most populated cities. However, as shown in Fig. 2e, there are significant disparities between Western and Eastern Europe, and Eastern European countries are abnormally distant. Thus, the African continent appears to be the second most distant. This also confirms the interpretation of the figure, which is that Oceania is abnormally close. To a lesser extent, this is also the case for South America.

Figure 9 presents the proportion of the five semantic nearest cities that are located in the same country (Fig. 9a and b) or continent (Fig. 10a and b). The first observation from the country-level analysis is that Asia (especially China), Western Europe, and North America have the highest rates. This suggests that the models have successfully captured geographical distances in their semantic representations for these regions. According to Tobler (1970)'s first law of geography (on page 236),

*Everything is related to everything else, but near things are more related than distant things*. Therefore, we observe the same inequalities that were highlighted previously, except

**Table 7** GDI ratio (semantic distance / geographic distance normalized by continent and model)

| Model | Metric | N. America | S. America | Europe | Africa | Asia | Oceania |
|---|---|---|---|---|---|---|---|
| BERT | mean | 1 | 0.99 | **1.11** | 1.09 | 1.07 | 0.88 |
| | farthest | 0 | 0 | **11** | 5 | 4 | 0 |
| | nearest | 6 | 4 | 0 | 0 | 4 | 6 |
| m-BERT | mean | 0.87 | 0.86 | **0.95** | 0.94 | 0.93 | 0.77 |
| | farthest | 1 | 0 | 6 | 6 | **7** | 0 |
| | nearest | 4 | 5 | 0 | 0 | 5 | 6 |
| RoBERTa | mean | 0.73 | 0.72 | **0.81** | 0.78 | 0.77 | 0.65 |
| | farthest | 0 | 0 | **15** | 0 | 5 | 0 |
| | nearest | 3 | 7 | 0 | 0 | 4 | 6 |
| GeoLM | mean | 0.85 | 0.84 | **0.96** | 0.92 | 0.9 | 0.75 |
| | farthest | 0 | 0 | **19** | 0 | 1 | 0 |
| | nearest | 5 | 4 | 0 | 0 | 5 | 6 |
| XLM-RoBERTa | mean | 0.7 | 0.69 | **0.78** | 0.75 | 0.74 | 0.62 |
| | farthest | 0 | 0 | **17** | 2 | 1 | 0 |
| | nearest | 3 | 7 | 0 | 0 | 4 | 6 |
| LLaMa 2 | mean | 0.96 | 0.95 | **1.07** | 1.04 | 1.03 | 0.86 |
| | farthest | 0 | 0 | **11** | 4 | 5 | 0 |
| | nearest | 5 | 5 | 0 | 0 | 4 | 6 |
| Mistral-v01 | mean | 0.98 | 0.96 | **1.09** | 1.05 | 1.04 | 0.88 |
| | farthest | 0 | 0 | **15** | 3 | 2 | 0 |
| | nearest | 4 | 6 | 0 | 0 | 4 | 6 |
| Openai/Ada | mean | 0.84 | 0.83 | **0.93** | 0.9 | 0.89 | 0.75 |
| | farthest | 0 | 0 | **18** | 1 | 1 | 0 |
| | nearest | 3 | 7 | 0 | 0 | 4 | 6 |
| country | | 17 | 12 | 35 | 47 | 38 | 6 |

**Mean**: the average GDI per continent. **Farthest**: the number of countries per continent among the 20 farthest (in **bold**). **Nearest**: the number of countries per continent among the 20 least distant (in underline)

*N. America* North America and *S. America* South America

for Oceania, which has low rates in this experiment because its cities are semantically closer to those of other continents, as suggested by the GDI ratio.

Unlike the previous indicators, the m-BERT models do not appear to improve upon BERT in these representations of distances. Undeniably, m-BERT has better geographical knowledge of underrepresented countries (i.e., countries for which their cities are not in the fixed vocabulary of the BERT-like model's tokenizer), but it does not capture geographical distance relationships any better.

The continent-level analysis, although yielding better scores than the country-level analysis, shows the same patterns of inequality.

## Geographic distribution of homonyms

The results of the second indicator (geographical knowledge quality) show that the American continent, particularly North America, experiences a performance decrease in our geo-probing task because of the large number of American cities that share names with European

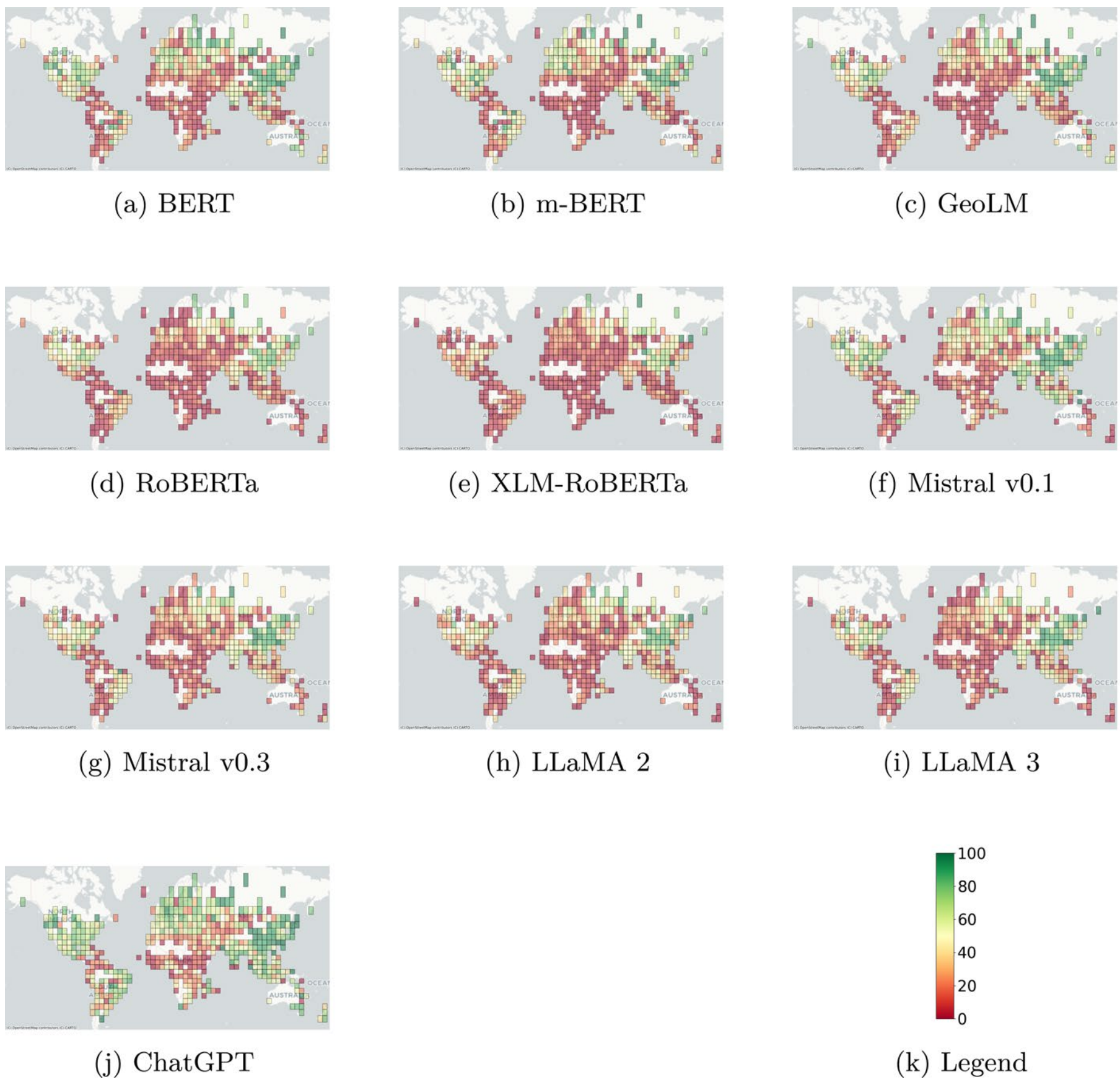

(a) BERT (b) m-BERT (c) GeoLM

(d) RoBERTa (e) XLM-RoBERTa (f) Mistral v0.1

(g) Mistral v0.3 (h) LLaMA 2 (i) LLaMA 3

(j) ChatGPT (k) Legend

**Fig. 9** Percentages of the 5 semantically closest cities in the same **country** by spatial aggregation in $5^\circ$ by $5^\circ$ pixels

(a) BERT (b) m-BERT (c) GeoLM

(d) RoBERTa (e) XLM-RoBERTa (f) Mistral v0.1

(g) Mistral v0.3 (h) LLaMA 2 (i) LLaMA 3

(j) ChatGPT (k) Legend

100
80
60
40
20
0

**Fig. 10** Percentages of the 5 semantically closest cities in the same **continent** by spatial aggregation in pixels of $5^{\circ}\times 5^{\circ}$

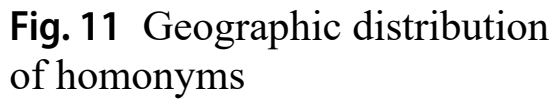

**Fig. 11** Geographic distribution of homonyms

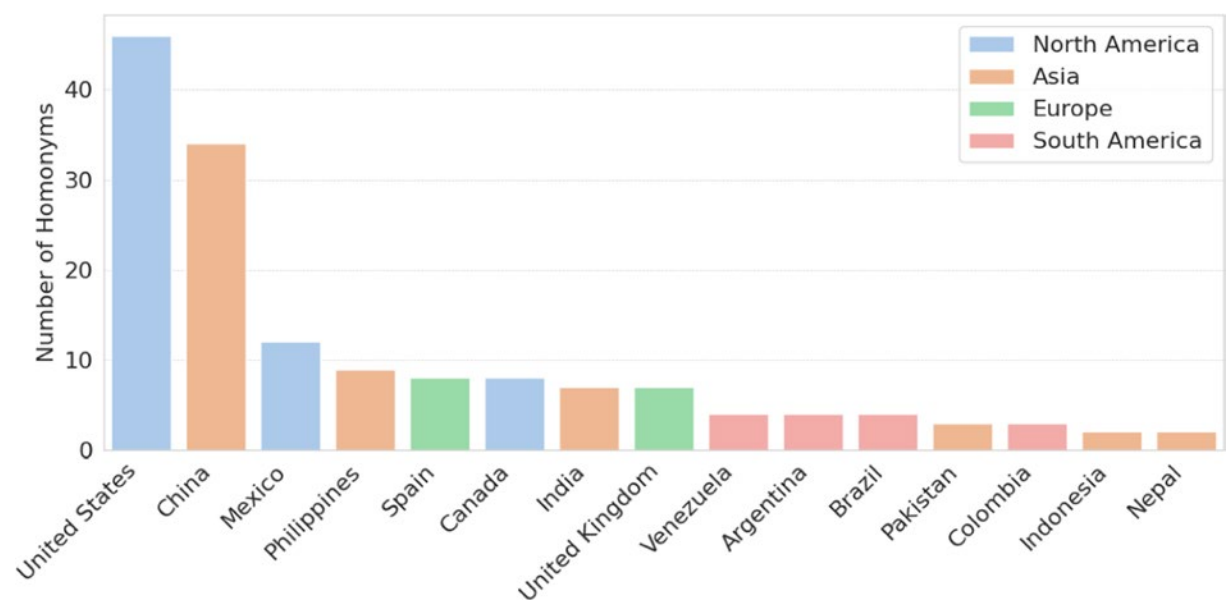

cities. These are homonyms of toponyms. Therefore, in this section, we analyse the distribution of toponym homonyms by continent.

To calculate this distribution, we rely on the list of cities worldwide with more than 100,000 inhabitants, and we count, for each country, the number of cities that have at least one homonym, which may, in some cases, be located within the same country. A visualization of the 15 countries with the highest number of homonyms is shown in Fig. 11. The United States, followed by China, has significantly more homonyms than all the other countries.

As mentioned in Sect. A.2, the high number of homonyms in the United States is explained by Zelinsky (1967)'s study, which highlights the tendency of 19th-century Americans to name new cities after older European ones. However, we do not have any hypothesis to explain the large number of homonyms in China.

**Supplementary Information** The online version contains supplementary material available at https://doi.org/10.1007/s10994-025-06916-9.

**Acknowledgements** This study was partially funded by EU grant 874850 MOOD. The contents of this publication are the sole responsibility of the authors and do not necessarily reflect the views of the European Commission.

**Data availability** The data used in this study are publicly available and can be accessed through the following sources: OpenDataSoft: Extraction of GeoNames of cities with more than 1,000 inhabitants. Available at: https://public.opendatasoft.com/explore/dataset/geonames-all-cities-with-a-population-1000/table/?flg=fr-fr&disjunctive.cou_name_en&sort=name Natural Earth Data: Get boundaries (polygon) of all countries. Available at: https://www.naturalearthdata.com/downloads/10m-cultural-vectors/ Kaggle World Capitals GPS. Available at: https://www.kaggle.com/datasets/nikitagrec/world-capitals-gps To reproduce figures and tables from this manuscript, we provide instructions through our GitHub repository at https://github.com/tetis-nlp/geographical-biases-in-llms/tree/master/paper-reproducibility. Note that a GPU with a minimum of 24 GB of RAM is needed. The total estimated execution time, if the indicators are run sequentially, is approximately 3 to 4 days. We also provide a lighter version of this pipeline that can be run inside a Google Colab environment. Four Python notebooks to illustrate the four indicators presented in this paper are available at https://github.com/tetis-nlp/geographical-biases-in-llms.